\documentclass[10pt,twocolumn,letterpaper]{article}

\usepackage{wacv}              

\definecolor{wacvblue}{rgb}{0.21,0.49,0.74}
\usepackage[pagebackref,breaklinks,colorlinks,allcolors=wacvblue]{hyperref}
\usepackage{multirow}
\usepackage{graphicx}
\usepackage{booktabs}
\usepackage{amsmath}
\usepackage{pifont}
\usepackage{array}
\usepackage{xcolor}
\definecolor{bestred}{RGB}{150,0,0} 
\definecolor{secondblue}{RGB}{0,45,150} 
\newcommand{\best}[1]{\textcolor{bestred}{#1}}
\newcommand{\second}[1]{\textcolor{secondblue}{#1}}

\newcommand{\cmark}{\ding{51}}
\newcommand{\xmark}{\ding{55}}
\def\wacvPaperID{641} 
\def\confName{WACV}
\def\confYear{2027}

\title{LSA-PAST: Layer-Specific Adaptation and Prototype Transport for Compositional Incremental Learning}

\vspace{-4mm}
\author{
Kyle Stein$^{1}$ \hspace{0.3em}
Andrew Arash Mahyari$^{2}$ \hspace{0.3em}
Guillermo Francia III$^{1}$ \hspace{0.3em}
Eman El-Sheikh$^{1}$ \hspace{0.3em}
Jiangpeng He$^{3}$ \\
$^{1}$University of West Florida, Pensacola, FL, USA\\
$^{2}$Florida Institute for Human and Machine Cognition, Pensacola, FL, USA\\
$^{3}$Indiana Univeristy, Bloomington, IN, USA\\
{\tt\small \{kstein,gfranciaiii,eelsheikh\}@uwf.edu, amahyari@ihmc.org,  jhe2@iu.edu}
}

\begin{document}
\maketitle
\begin{abstract}
Class-Incremental Learning (CIL) enables models to recognize new classes over time while retaining prior knowledge, but many real-world problems are compositional, where visual concepts arise from combinations of primitives such as states (i.e., red, shiny) and objects (i.e., shoe, car). Standard CIL methods treat each composition as an independent class, ignoring shared primitives and providing no mechanism to stabilize their representations as new compositions arrive. This causes parameter updates to entangle primitives and induce drift that degrades performance on earlier compositions.  Compositional Incremental Learning (composition-IL) formalizes this setting by introducing new state–object compositions across tasks. In this paper, we propose Layer-Specific Adaptation with Prototype-Anchored Adapter-Space Transport (\textbf{LSA-PAST}), a parameter-efficient method tailored to composition-IL. LSA integrates Low-Rank Adaptation (LoRA) into a frozen ViT with depth-specialized adapters for states, objects, and compositions. PAST then maps past, anchored prototypes into the current adapter space through a rehearsal-free affine transform fit on current-task data, enabling old prototypes to remain compatible with new adapters. Extensive experiments demonstrate that LSA-PAST achieves state-of-the-art (SOA) accuracy-forgetting trade-off performance for composition-IL.
\end{abstract}
    
\vspace{-6mm}
\section{Introduction}
\label{sec:intro}
\vspace{-2mm}

Deep learning has achieved remarkable success across visual recognition tasks, from fine-grained identification to large-scale categorization \cite{he2016deep, li2022mvitv2, boutros2022elasticface}. This success typically rests on the brittle assumption that the world can be captured in a single, static training set. However, in real-world deployments, data arrives sequentially and models must incorporate new knowledge without full retraining or access to past data. Class-Incremental Learning (CIL) \cite{kirkpatrick2017overcoming, li2017learning, rebuffi2017icarl, liu2025class, meng2024diffclass} formalizes this setting where a learner is exposed to a sequence of tasks with expanding label space and must acquire new classes while preserving previously learned knowledge, reinforcing the issue of catastrophic forgetting \cite{french1999catastrophic}. Recent CIL methods adopt parameter-efficient fine-tuning (PEFT) of pre-trained models to alleviate forgetting through prompt-tuning \cite{wang2022learning,smith2023coda, wang2022dualprompt} and adapters \cite{zhou2024expandable, he2025cl, li2025dynamic}. However, this progress is largely measured in object-only settings where classes are treated as singular entities, an assumption misaligned with the compositional nature of real-world concepts.

\begin{figure}[t!]
    \centering
    \includegraphics[width=1.0\columnwidth]{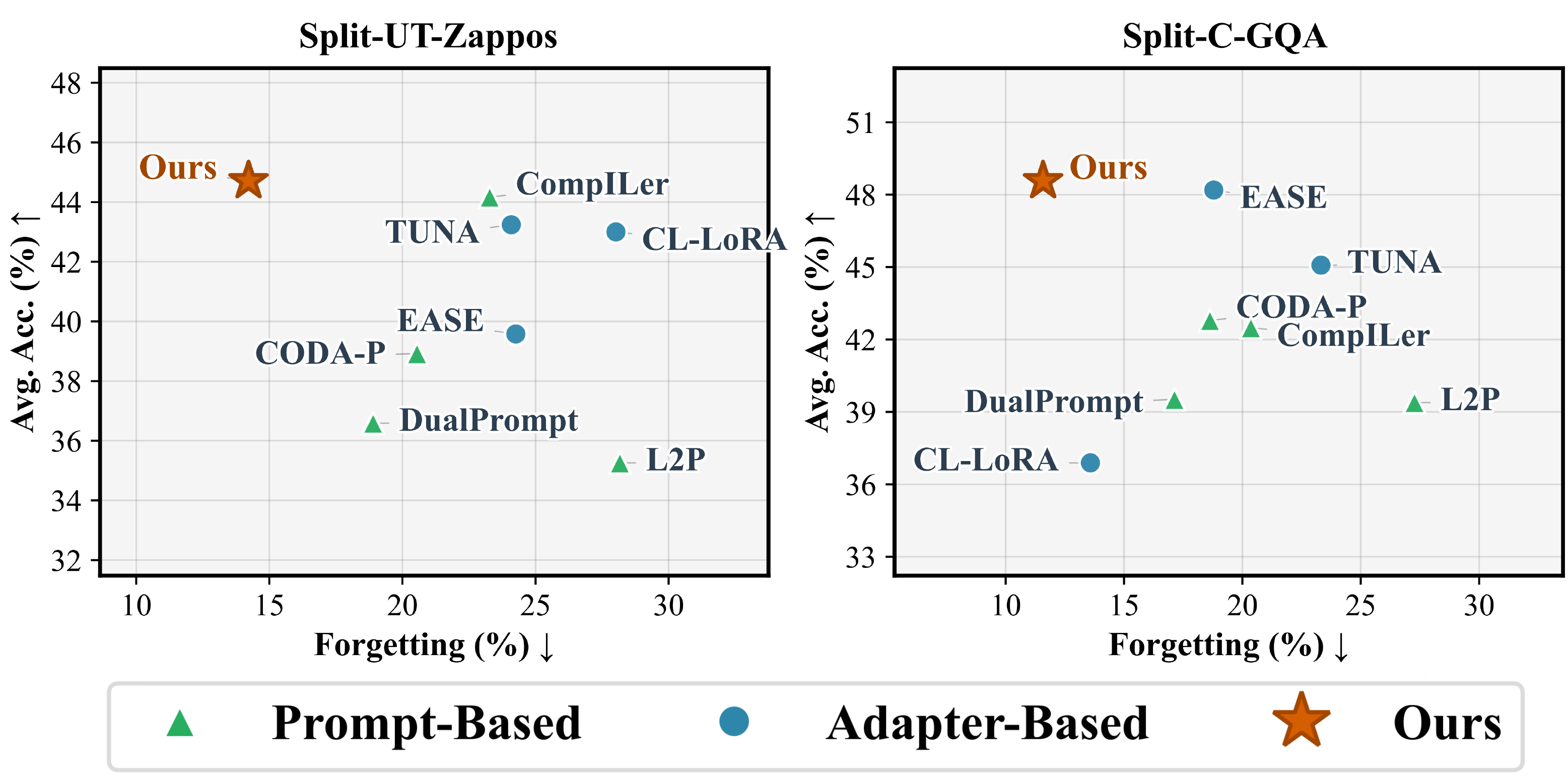}
    \caption{Accuracy–forgetting trade-off on composition-IL benchmarks for adapter- and prompt-based methods.}
    \label{fig:tradeoff}
    \vspace{-6mm}
\end{figure}

Concepts in the visual world are often \emph{compositional}, combining primitives such as states (e.g., \emph{striped}, \emph{leather}) and objects (e.g., \emph{shirt}, \emph{shoe}) \cite{lake2014towards, lake2017building, frankland2020concepts}. This structure is central to human cognition, yet most CIL methods and benchmarks treat classes as atomic and ignore compositionality. Compositional Incremental Learning (composition-IL), introduced in \cite{zhang2024not}, addresses this gap by learning to acquire new state–object pairs over tasks while retaining performance on prior compositions. A key difficulty in composition-IL is \emph{state confusion}: models often recognize the object but confuse its state (e.g., ``shoe'' vs. ``leather shoe'' or ``canvas shoe''). Pretrained models bias representations toward object identity \cite{geirhos2018imagenet, naseer2021intriguing}, causing feature entanglement among compositions sharing an object \cite{hao2023learning, lu2023decomposed}. Furthermore, standard pretraining often lacks the fine-grained sensitivity to texture and material required for state discrimination \cite{geirhos2018imagenet, wei2019adversarial}. These issues worsen when object-centric CIL methods are applied to compositions. Fig.~\ref{fig:tradeoff} shows that prior, object-centric CIL methods yield poor accuracy-forgetting trade-offs, while our method provides the strongest overall balance across all benchmarks.

In this work, we propose Layer-Specific Adaptation with Prototype-Anchored Adapter-Space Transport (LSA-PAST) for composition-IL. Our approach introduces \textit{Layer-Specific Adaptation} (LSA) with three depth-specialized Low-Rank Adaptation (LoRA) \cite{hu2022lora} adapters that exploit the hierarchical processing structure of vision transformers (ViTs) \cite{dosovitskiy2020image, dorszewski2025colors, NEURIPS2024_6216515a, lepori2024beyond, raghu2021vision}: a state adapter in early layers reduces state confusion by amplifying low-level cues such as texture and color, an object adapter in middle layers learns shape and part structure, and a composition adapter in late layers integrates primitives into complete concepts. By exploiting this property and specializing transformer depths for different aspects of compositional reasoning, the model learns primitive-specific representations that transfer across compositional classes.

To improve the stability of compositional decision boundaries, we introduce \emph{Primitive-Guided Stabilization} (PGS), which anchors each composition prototype to directions from the state and object heads when its primitives recur across tasks. Finally, to alleviate catastrophic forgetting arising from inter-task prototype misalignment, we propose \emph{Adapter–Space Transport} (AST). In the composition-IL setting, each task has unique LoRA adapters, so features and prototypes learned at different times live in task-specific adapter spaces. These task-specific adapter spaces are not directly interoperable, limiting the use of cross-space evidence for previously learned compositions \cite{yu2020semantic}. AST addresses this by fitting a lightweight affine relation from each past adapter space to the \emph{current} one using only \emph{current-task} image pairs (rehearsal-free), and transporting the anchored unit prototypes from past tasks into the current adapter geometry. Overall, our key contributions can be summarized as follows:

\vspace{-2mm}
\begin{itemize}
    \item We introduce \textbf{LSA-PAST}, a parameter-efficient composition-IL framework with novel depth-specialized adaptation and rehearsal-free prototype transport across task-specific adapter spaces.
    \item We curate a new composition-IL benchmark for evaluating novel state--object combinations and recurring primitives across tasks.
    \item We provide strong baselines, ablations, and analysis showing improved accuracy-forgetting trade-offs on composition-IL benchmarks.
\end{itemize}

\vspace{-4mm}
\vspace{-1mm}
\section{Related Work}
\label{sec:related_work}
\vspace{-2mm}
\subsection{Class-Incremental Learning}
\vspace{-1mm}
\noindent \textbf{Class-Incremental Learning (CIL)} aims to learn new classes while retaining prior knowledge \cite{zhou2024class}. Classical methods such as EWC, LwF, and iCaRL respectively regularize parameter changes, distill from previous models, or combine exemplar replay with nearest-mean classification \cite{kirkpatrick2017overcoming, li2017learning, rebuffi2017icarl}. Exemplar-free CIL avoids storing past samples by regularizing the current model or allocating new capacity per task \cite{zhou2024class, he2025cl}, but still ignores compositional structure. Beyond the standard setting where tasks introduce disjoint class sets, related fields include multi-label CIL (MLCIL) \cite{song2024overcoming, dong2023knowledge}, which handles multi-label predictions per image, and blurry-CIL \cite{bang2021rainbow, moon2023online}, which allows overlapping class distributions across tasks. Related continual compositional works study continual compositional zero-shot learning (CCZSL) \cite{continualczsl,maryam2025prompt} aimed at progressively improving zero-shot compositional generalization \cite{li2026composition}. Composition-IL instead focuses on learning new state--object compositions, even when states and objects recur across tasks.

\noindent \textbf{CIL with pre-trained models} has shifted from training from scratch to adapting frozen pre-trained models (PTMs) \cite{dosovitskiy2020image} with parameter-efficient methods \cite{wang2022learning, zhou2024continual}. PTMs encode rich visual knowledge and constrain plasticity to a low-dimensional subspace, reducing representation drift across tasks. We therefore adapt PTMs for composition-IL with parameter-efficient fine-tuning (PEFT), broadly categorized into prompt-based and adapter-based approaches.

\noindent \textbf{Prompt-based methods} introduce learnable tokens that steer a frozen backbone with high parameter efficiency \cite{lester2021power, jia2022visual}. L2P retrieves prompts through query-key matching \cite{wang2022learning}, Dual-Prompt separates shared and task-specific prompts \cite{wang2022dualprompt}, CODA-Prompt composes multiple prompts without full fine-tuning \cite{smith2023coda}, and CompILer \cite{zhang2024not} extends this idea to composition-IL with multiple L2P-style prompt pools routed into state, object, and composition heads. Despite their efficiency, prompt-based methods often converge more slowly and are more difficult to optimize than adapter-based methods \cite{ding2023parameter}.

\noindent \textbf{Adapter-based methods} inject lightweight modules into transformer blocks to shape intermediate features, mitigating forgetting by localizing updates to specific subspaces \cite{han2024parameter}. Notable methods include EASE \cite{zhou2024expandable}, which adds task-specific expandable modules, CL-LoRA \cite{he2025cl}, which employs a dual-adapter design, and TUNA \cite{wang2025integrating}, which combines entropy-guided task-specific adapters with a fused universal adapter. However, these methods typically treat classes as independent entities, ignoring the compositional structure of images. Our method addresses this gap through depth specialization and cross-subspace alignment to stabilize novel compositional pairs.

\vspace{-2mm}
\subsection{Hierarchical Processing in Vision Transformers}
\vspace{-1mm}
Recent studies reveal that ViTs \cite{dosovitskiy2020image} process visual information through distinct stages of integration across depth \cite{dorszewski2025colors, raghu2021vision, NEURIPS2024_6216515a, lepori2024beyond}. \emph{Early layers} (e.g., 1-4) primarily extract disentangled primitive attributes such as color, texture, and simple edges. These layers exhibit highly localized receptive fields, resembling early convolutional filters that capture low-level statistics and fine-grained appearance cues \cite{raghu2021vision, dorszewski2025colors}. \emph{Middle layers} (e.g., 5-8) progressively integrate these primitives into structured, object-level representations. As receptive fields expand, attention begins to aggregate features across parts, capturing relationships between components that define an object’s shape and spatial layout \cite{lepori2024beyond, NEURIPS2024_6216515a}. \emph{Late layers} perform global  operations over integrated representation, where  attention heads attend globally across the entire image \cite{raghu2021vision}, capturing contextual relationships between objects and implementing operations that generalize to novel feature combinations \cite{NEURIPS2024_6216515a, lepori2024beyond}. This progression motivates our depth-specialized design. Unlike uniform approaches \cite{he2025cl}, we align adapter placement within stages of the ViT layers based on this evidence. 

\vspace{-3mm}

\section{Preliminaries}
\label{sec:preliminaries}

\begin{figure*}[t!]
\centering
\includegraphics[width=0.9\linewidth,trim={0cm 0cm 0cm 0cm},clip]{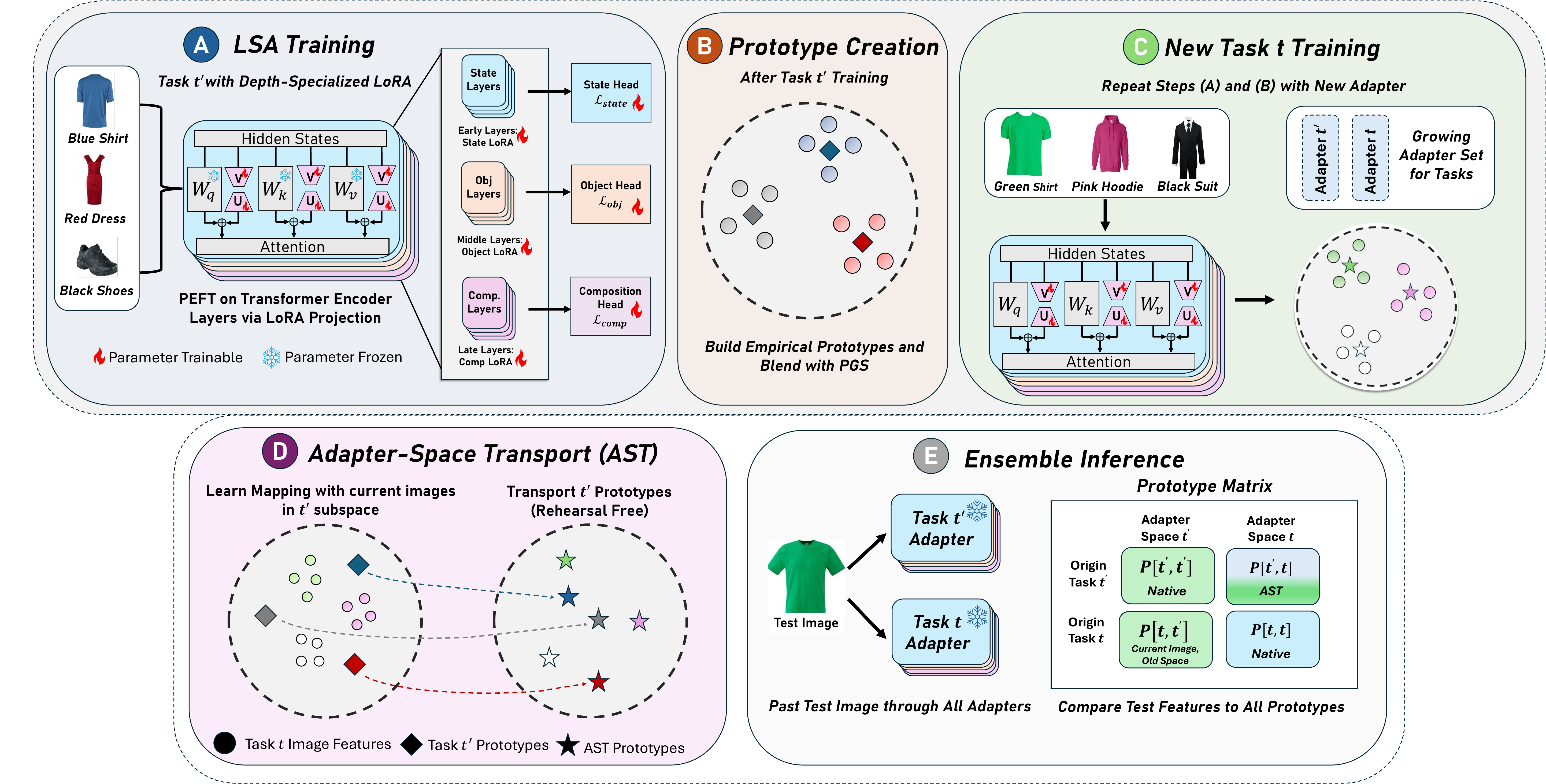}
\caption{\textbf{(A)} We adapt a frozen ViT with depth-specialized LoRA feeding three heads. \textbf{(B)} Post-training, empirical prototypes are blended with primitive directions (PGS) and stored. \textbf{(C)} Subsequent tasks instantiate fresh adapters for new compositions. \textbf{(D)} Using current-task images as transient correspondences, AST fits an affine map $\phi_{t'\rightarrow t}$ to transport past prototypes into the current adapter space without rehearsal. \textbf{(E)} Inference ensembles cosine similarities across all adapter spaces,  recognizing both new and previously learned compositions.}
\label{fig:Main_Fig}
\vspace{-4mm}
\end{figure*}

\vspace{-1mm}
\subsection{Problem Setup}
In composition-IL, visual concepts of images are expressed as compositions of two primitive components: the \textit{state} and an \textit{object}. Let the state set be $\mathcal{S}=\{s_1, s_2, \dots, s_n\}$ and the object set be $\mathcal{O}=\{o_1, o_2, \dots, o_m\}$. Each composition class is defined as $c=\langle s, o \rangle$, where $s \in \mathcal{S}$ and $o \in \mathcal{O}$, forming the global set of possible compositions $\mathcal{C} \subseteq \mathcal{S} \times \mathcal{O}$. Learning proceeds over a sequence of $N$ tasks $\mathcal{T} = \{\mathcal{T}_1, \mathcal{T}_2, \dots, \mathcal{T}_N\}$, where each task $\mathcal{T}_t$ introduces a disjoint set of composition classes $\mathcal{C}^{(t)} \subset \mathcal{C}$, such that $\mathcal{C}^{(i)} \cap \mathcal{C}^{(j)} = \emptyset$ for any $i \neq j$. Unlike compositions, the primitive classes (states and objects) may recur across tasks, allowing knowledge of previously seen primitives to benefit new compositions. Each task $\mathcal{T}_t$ provides a labeled dataset $\mathcal{D}^{(t)} = \{(x_i, y_i)\}$, where each label $y_i = \langle s_i, o_i \rangle$ denotes the composition of state $s_i$ and object $o_i$. Furthermore, we follow the challenging \textbf{rehearsal-free setting}, where no exemplars from previous tasks are leveraged for knowledge replay \cite{fukuda2025adapter, zhou2024expandable, smith2023coda}. The model must learn to recognize the new compositions while maintaining performance on all previously seen compositions.

\subsection{Adapter-Based Composition-IL}
Following prior work \cite{zhang2024not, zhou2024expandable, he2025cl, fukuda2025adapter}, we adopt a pretrained ViT \cite{dosovitskiy2020image} and perform adapter-based incremental learning with LoRA \cite{hu2022lora}. In this section, we introduce main components of our method in forming task-specific adapter spaces with LoRA and prototype-classifier for composition-IL.

\noindent \textbf{Low-Rank Adaptation:} LoRA \cite{hu2022lora} enables PEFT of pre-trained models by freezing the backbone weights and injecting trainable low-rank adapters into the attention projections of transformer layers. For a pre-trained weight matrix $\mathbf{W_0} \in \mathbb{R}^{d \times k}$, LoRA modifies the feature extraction as:

\vspace{-1mm}
\begin{equation}
\mathbf{y} = \mathbf{W_0}\mathbf{x} + \Delta\mathbf{W}\,\mathbf{x}, 
\label{eq:lora}
\end{equation}

\noindent where $\mathbf{x}$ and $\mathbf{y}$ denote the input and output activations, respectively, and $\Delta\mathbf{W}=\mathbf{U}\mathbf{V}$ is a low-rank adaptation. Here, $\mathbf{V}\!\in\!\mathbb{R}^{r\times k}$ down-projects into a rank-$r$ subspace and $\mathbf{U}\!\in\!\mathbb{R}^{d\times r}$ projects back to the output. Only $(\mathbf{U},\mathbf{V})$ are trainable, adding $r(d{+}k)$ parameters while retaining the pretrained backbone. We apply LoRA to the Q/K/V projection matrices of the pre-trained ViT \cite{dosovitskiy2020image}. To reduce cross-task interference while preserving parameter efficiency, we instantiate \emph{one LoRA adapter per task}: for task $t$ we optimize only the rank-$r$ updates $\{\Delta\mathbf{W}^{(t)}\}$ in selected layers, while all prior adapters $\{\Delta\mathbf{W}^{(t')}\}_{t'<t}$ remain frozen.

\noindent \textbf{Prototype-Classifier for Composition-IL:}
Standard CIL methods \cite{zhou2024expandable, he2025cl, wang2025integrating} pair the backbone with a single linear head trained with cross-entropy on each task. In our depth-specialized setting for composition-IL, we instead use three heads during training (state, object, composition), but at the end of task $t$, only the composition head is replaced with a prototype classifier built from task-$t$ features. Let $f_{\text{comp}}^{(t)}(x) \in \mathbb{R}^d$ be the composition feature from the final ViT layer within the task-$t$ adapter space, and
$\widehat z_{\text{comp}}^{(t)}(x) = \operatorname{unit}\!\big(f_{\text{comp}}^{(t)}(x)\big)$ its L2-normalized version, where we define $\mathrm{unit}(\mathbf{v}) = \frac{\mathbf{v}}{\|\mathbf{v}\|_2}$. For each composition class $c \in \mathcal{C}^{(t)}$, we compute the empirical prototype as:

\vspace{-4mm}
\begin{equation}
p_c^{\text{emp},(t)} \;=\; \operatorname{unit}\!\Big(\frac{1}{|\mathcal{D}^{(t)}(c)|}\sum_{x_i\in\mathcal{D}^{(t)}(c)} \widehat z_{\text{comp}}^{(t)}(x_i)\Big),
\label{eq:prototype}
\end{equation}
\vspace{-4mm}

\noindent where $\mathcal{D}^{(t)}(c)=\{x_i \in \mathcal{D}^{(t)} : y_i=c\}$. The resulting prototypes form the composition head used for inference, while the state and object heads remain standard linear classifiers for auxiliary supervision.

\vspace{-1mm}
\section{Methodology}
\label{sec:methodology}
\vspace{-1mm}
\subsection{Overview}
LSA-PAST combines three components for composition-IL. First, LSA attaches task-specific LoRA adapters to different ViT depths for state, object, and composition reasoning. Second, PGS anchors empirical composition prototypes using recurring state and object directions. Third, AST maps old prototypes into new task adapter spaces through a rehearsal-free affine transform fitted only on current-task images. Together, depth-specialized LoRA and prototype-anchored transport balance \emph{stability} from frozen adapters with \emph{plasticity} for new compositions (Fig.~\ref{fig:Main_Fig}).

\subsection{Layer-Specific Adaptation with LoRA}
\label{sec:lora}
We introduce \textbf{LSA}, with LoRA specialization modules applied at selected depths of a ViT. Motivated by evidence that ViT representations evolve with depth \cite{dorszewski2025colors, NEURIPS2024_6216515a, lepori2024beyond, raghu2021vision}, we allocate separate LoRA branches for states, objects, and compositions to different depth ranges instead of adapting all layers uniformly. This depth specialization exploits the ViT hierarchy and allocates adaptation capacity only where needed, providing finer parameter control than uniform LoRA. In the incremental setting, we instantiate a \textbf{\textit{separate set of adapters per task}}; once a task finishes, its adapters are frozen and a fresh set is created for the subsequent task.

\noindent \textbf{LoRA-Branch Configuration:} Let $\mathbf{W}_0 \in \mathbb{R}^{d \times k}$ denote a frozen attention projection matrix (query, key, or value) in the ViT backbone. For a given layer $\ell$, we augment $\mathbf{W}_0$ with at most one low-rank adaptation, $\Delta\mathbf{W}$ (Eq.\ref{eq:lora}). Rather than applying uniform LoRA across all layers, we exploit the hierarchical processing structure of ViTs. We partition the encoder into three disjoint depth ranges aligned with this hierarchy. This design helps mitigate state confusion by allocating early-layer adaptation to state cues such as texture and color, which are often suppressed by object-dominant representations in deeper layers. For layer $\ell$ assigned to branch $b \in \{\text{state}, \text{obj}, \text{comp}\}$, the depth-specialized adaptation becomes:

\vspace{-2mm}
\begin{equation}
\Delta\mathbf{W}_\ell^b = \mathbf{U}_\ell^b \mathbf{V}_\ell^b,
\label{eq:depth_lora}
\end{equation}

\noindent where $\mathbf{U}_\ell^b \in \mathbb{R}^{d \times r_b}$ and $\mathbf{V}_\ell^b \in \mathbb{R}^{r_b \times k}$ are trainable with $r_b \ll \min(d,k)$. This configuration enables fine-grained control: each branch can use different ranks $(r_{\text{state}}, r_{\text{obj}}, r_{\text{comp}})$ while maintaining parameter efficiency through non-overlapping layer assignments. During training on task $t$, only the adapters for the current task $\{\mathbf{U}_\ell^{b,t},\mathbf{V}_\ell^{b,t}\}$ are updated and all previous task adapters $\{\mathbf{U}_\ell^{b,t'},\mathbf{V}_\ell^{b,t'}\}_{t' < t}$ remain frozen.

\noindent \textbf{Multi-Branch Training Supervision:} To leverage the depth-specialized LoRA adaptations, we introduce a multi-branch training scheme with dedicated classification heads for each semantic level. We maintain three linear classification heads that are updated incrementally across tasks, each parameterized by a weight matrix: $\mathbf{W}_{\text{state}} \in \mathbb{R}^{|\mathcal{S}| \times d}$, $\mathbf{W}_{\text{obj}} \in \mathbb{R}^{|\mathcal{O}| \times d}$, and $\mathbf{W}_{\text{comp}} \in \mathbb{R}^{|\mathcal{C}| \times d}$, where $\mathcal{S}$, $\mathcal{O}$, and $\mathcal{C}$ denote the sets of states, objects, and compositions, respectively. Each head is trained using cross entropy loss:

\vspace{-3mm}
\begin{equation} 
\resizebox{0.90\columnwidth}{!}{$\displaystyle \mathcal{L}_{\text{comp}}^{(t)}(x,y) \;=\; -\log \frac{ \exp\!\big([\mathbf{W}_{\text{comp}} \,\widehat z_{\text{comp}}^{(t)}(x)]_{y}\big) }{ \sum_{c \in \mathcal{C}^{(t)}} \exp\!\big([\mathbf{W}_{\text{comp}} \,\widehat z_{\text{comp}}^{(t)}(x)]_{c}\big) } $} 
\end{equation}

\noindent where $[\cdot]_c$ denotes the $c$-th element of the logit vector. The losses $\mathcal{L}_{\text{state}}$ and $\mathcal{L}_{\text{obj}}$ are defined analogously. During training, the weighted multi-task loss backpropagates gradients through only the \emph{current task's} adapters. At inference, linear composition head is replaced with prototype-based classification (Eq.~\ref{eq:prototype}), using prototypes maintained across tasks.

\subsection{Prototype Anchoring with Primitive-Guided Stabilization}
\label{sec:PGS}
Composition-IL naturally recurs primitives across tasks. For example, after learning ``Red Shoe'' in Task 1, Task 2 might introduce ``Red Hat'' (recurring state) or ``Blue Shoe'' (recurring object). PGS leverages this recurrence by blending empirical prototypes with primitive-guided directions at the end of each task. This provides a recurrence-dependent stabilization mechanism while preserving the task-native empirical representation.


\noindent\textbf{Blend prototypes with the final primitive direction:}
After training task $t$, for each composition class introduced during that task, we first build empirical prototypes by averaging unit-normalized features from the training samples per compositional class (and re-normalizing) to obtain $p_c^{\text{emp},(t)}$ (Eq.~\ref{eq:prototype}). We then construct the end-of-task PGS direction from the primitive head weights after training task $t$. Let $\mathbf{w}_s$ and $\mathbf{w}_o$ denote the weight vectors corresponding to state $s$ and object $o$:

\begin{equation}
\mathbf{d}^{\text{end}}_{\text{PGS}}(c) = \operatorname{unit}\left(\operatorname{unit}(\mathbf{w}_s) + \operatorname{unit}(\mathbf{w}_o)\right).
\label{eq:d-end}
\end{equation}

\noindent Although $\mathbf{w}_s$ and $\mathbf{w}_o$ arise from their respective depth-specific state and object representations, PGS uses their normalized classifier weights as discriminative primitive directions rather than empirical feature prototypes, consistent with prior work demonstrating that normalized classifier weights encode meaningful class directions in representation space \cite{qi2018low}. PGS uses these directions as weak cross-depth primitive guidance rather than assuming identical feature geometry. To stabilize prototypes across tasks, we lightly blend $p_c^{\text{emp},(t)}$ with $\mathbf{d}^{\text{end}}_{\text{PGS}}(c)$ using a recurrence-weighted factor $\eta_c$ to obtain the final native prototype in the task-$t$ adapter space:

\vspace{-2mm}
\begin{equation}
p_c^{(t)} \leftarrow \operatorname{unit}\!\big((1-\eta_c)\,p_c^{\text{emp},(t)} + \eta_c\, \mathbf{d}_{\text{PGS}}^{\text{end}}(c)\big).
\label{eq:p_c}
\end{equation}

\noindent Blending is applied \emph{once}, when the compositional class first appears and prototypes from earlier tasks are not re-blended thereafter. Furthermore, the blending weight $\eta_c$ is determined by primitive recurrence at the time the composition was introduced. Let $\mathcal{S}^{<t}$ and $\mathcal{O}^{<t}$ denote states and objects seen before task $t$. For each composition $c=\langle s,o\rangle$, we set:

\begin{equation}
\eta_c =
\begin{cases}
\eta_{\text{both}}, &
\text{if } s \in \mathcal{S}^{<t} \text{ and } o \in \mathcal{O}^{<t}, \\[4pt]
\eta_{\text{one}}, &
\text{if } (s \in \mathcal{S}^{<t}) \oplus (o \in \mathcal{O}^{<t}), \\[4pt]
0, &
\text{if } s \notin \mathcal{S}^{<t} \text{ and } o \notin \mathcal{O}^{<t}.
\end{cases}
\label{eq:eta_c}
\end{equation}

When neither primitive has been seen ($\eta_c = 0$), $p_c^{(t)}$ remains equal to the empirical prototype $p_c^{\text{emp},(t)}$. When primitives recur, the state and object classifiers retain and refine their weight vectors across tasks, providing reusable discriminative guidance for blending. Therefore, PGS anchors new composition prototypes using learned primitive directions to improve stability across tasks.

\subsection{Adapter-Space Transport}
\label{sec:proto_mapping}
Each task $\mathcal{T}_t$ instantiates its own LoRA adapters (Sec.~\ref{sec:lora}), which are frozen once training completes. Consequently, composition embeddings from different tasks live in \emph{adapter-specific} feature spaces. We refer to the feature space induced by the adapters trained for task $a$ as \emph{adapter space $a$}. At the end of task $t$, we maintain a block-structured prototype matrix $P$, where each block $P[t',a]\in\mathbb{R}^{|\mathcal{C}^{(t')}|\times d}$ stores unit prototypes for classes introduced at task $t'$ but expressed in adapter space $a \le t$. Diagonal blocks contain native PGS prototypes, while off-diagonal blocks store cross-subspace prototypes.

For a new task $t$, we first fill $P[t,t]$ with PGS-anchored prototypes from current-task images under the task-$t$ adapters. For the same classes, we obtain $P[t,a]$ for each earlier adapter space $a<t$ by passing current-task images through frozen past adapters. In contrast, for classes introduced in earlier tasks $t'<t$, we lack their images in the rehearsal-free setting \cite{zhou2024expandable, he2022exemplar, he2025cl} and cannot directly compute their prototypes under the current adapters. Their native prototypes remain reliable in their originating adapter spaces, but native-space scoring leaves the task-specific spaces isolated. AST instead makes these spaces interoperable by using current-task images as transient correspondences to express prior classes in later adapter spaces without rehearsal.


\vspace{0.5mm}
\noindent \textbf{Cross-Subspace Alignment:}
For a current task $t$ and any previous task index $t' < t$, we take the same current-task image $x \in \mathcal{D}^{(t)}$ and extract L2-normalized composition embeddings under both adapter sets: $\widehat z_{\text{comp}}^{(t')}(x) \in \mathbb{R}^d$ from the frozen task-$t'$ adapters and $\widehat z_{\text{comp}}^{(t)}(x) \in \mathbb{R}^{d}$ from the current adapters. The pairs $\big(\widehat z_{\text{comp}}^{(t')}(x), \widehat z_{\text{comp}}^{(t)}(x)\big)$ provide transient correspondences that identify a cross-subspace relation between adapter spaces \cite{gomez2024exemplar,fukuda2025adapter,yu2020semantic,he2025semantic,wu2025navigating}. Using only these current-task correspondences, we fit a regularized affine map directly from each past adapter space $t'$ to the current space $t$ and apply it to prior prototypes without rehearsal. By default, we use all current-task images and vary this number in our transport-budget ablation.

We define the adapter-space transport map $\phi_{t'\!\rightarrow t}:\mathbb{R}^d\!\rightarrow\!\mathbb{R}^d$ as an affine transformation:

\vspace{-1mm}
\begin{equation}
\phi_{t'\!\rightarrow t}(u)\;=\;\mathbf{A}_{t'\!\rightarrow t}\,u\;+\;\mathbf{b}_{t'\!\rightarrow t},
\end{equation}

\noindent where $\mathbf{A}_{t'\!\rightarrow t}\in\mathbb{R}^{d\times d}$ and $\mathbf{b}_{t'\!\rightarrow t}\in\mathbb{R}^{d}$. For a fixed past adapter index $t' < t$, we estimate $(\mathbf{A},\mathbf{b})\equiv(\mathbf{A}_{t'\!\rightarrow t},\mathbf{b}_{t'\!\rightarrow t})$ at the end of task $t$ by minimizing:

{\small
\begin{equation}
\min_{\mathbf{A},\mathbf{b}}\; \sum_{x_i \in \mathcal{D}^{(t)}} \big\| \widehat z_{\text{comp}}^{(t)}(x_i) -
\big(\mathbf{A}\, \widehat z_{\text{comp}}^{(t')}(x_i) + \mathbf{b}\big) \big\|_2^2 \;+\; \lambda \|\mathbf{A}\|_F^2,
\label{eq:ast_regression}
\end{equation}
}

\noindent where the ridge term $\lambda\|\mathbf{A}\|_F^2$ regularizes the linear component while leaving the bias unconstrained. With unit-normalized features, minimizing this objective aligns old-adapter directions to current-adapter directions on the current-task distribution. We fit $\phi_{t'\!\rightarrow t}$ once per pair $(t',t)$ at the end of task $t$ only into the current adapter space and \textit{do not refit maps between earlier adapters}. 

\noindent \textbf{Prototype Transport:}
After fitting maps for all $t' < t$, we transport all previously seen prototypes into the current adapter space. At the end of task $t$, we already have the native prototype $p_c^{(t)}$ for each new class $c \in \mathcal{C}^{(t)}$ from PGS (Eq.~\ref{eq:p_c}), stored as the native diagonal block $P[t,t]$ of the prototype matrix, and we have directly computed cross-subspace prototypes $P[t,a]$ for all $a < t$ by passing current-task images through older adapters. More generally, for any composition class $c$ introduced at task $t_c$, its stored, blended prototype $p_c^{(t_c)}$ remains in its originating adapter space as the diagonal block $P[t_c,t_c]$.

Applying $\phi_{t'\!\rightarrow t}$ to these native prototypes yields a transported view in the current adapter space (adapter space $t$):

\begin{equation}
\tilde p_c^{(t'\!\rightarrow t)} \;=\; \operatorname{unit}\!\big(\mathbf{A}_{t'\!\rightarrow t}\, p_c^{(t')} + \mathbf{b}_{t'\!\rightarrow t}\big),
\end{equation}

\noindent for all classes $c$ whose origin task is $t' = t_c < t$. The transported prototypes $\tilde p_c^{(t'\!\rightarrow t)}$ are stored in the above-diagonal block $P[t',t]$. Intuitively, $\phi_{t'\!\rightarrow t}$ determines how old classes would be embedded by the current adapters, based on how current-task images move between adapter spaces. This procedure is performed after every task: at the end of task $t$ we build new native prototypes $P[t,t]$ and their cross-subspace views $P[t,a]$ for all $a < t$ using current images, and fit a map $\phi_{t'\!\rightarrow t}$ for each $t'<t$ and use it to update the column $P[t',t]$. No past-task images are stored or replayed, the only persisted state comprises frozen adapters, PGS-anchored prototypes, and their transported versions.

\begin{table*}[t!]
  \centering
  \small
  \setlength{\tabcolsep}{2.2pt}
  \resizebox{\textwidth}{!}{%
    \begin{tabular}{l l c cc cc cc cc}
      \toprule
      \multirow{3}{*}{Method type} & \multirow{3}{*}{Method} & \multirow{3}{*}{\%Params}
        & \multicolumn{2}{c}{Split-Clothing}
        & \multicolumn{4}{c}{Split-UT-Zappos}
        & \multicolumn{2}{c}{Split-C-GQA} \\
      \cmidrule(lr){4-5}\cmidrule(lr){6-9}\cmidrule(lr){10-11}
      & & &
        \multicolumn{2}{c}{5 tasks}
        & \multicolumn{2}{c}{5 tasks}
        & \multicolumn{2}{c}{10 tasks}
        & \multicolumn{2}{c}{5 tasks} \\
      \cmidrule(lr){4-5}\cmidrule(lr){6-7}\cmidrule(lr){8-9}\cmidrule(lr){10-11}
      & & &
        Avg Acc \% ($\uparrow$) & FTT \% ($\downarrow$)
        & Avg Acc \% ($\uparrow$) & FTT \% ($\downarrow$)
        & Avg Acc \% ($\uparrow$) & FTT \% ($\downarrow$)
        & Avg Acc \% ($\uparrow$) & FTT \% ($\downarrow$) \\
      \midrule

      \multirow{3}{*}{Classic CIL}
        & EWC \cite{kirkpatrick2017overcoming} & 100
        & $47.89\text{\scriptsize$\pm0.87$}$ & $52.75\text{\scriptsize$\pm0.44$}$
        & $37.59\text{\scriptsize$\pm2.06$}$ & $55.70\text{\scriptsize$\pm2.76$}$
        & $24.63\text{\scriptsize$\pm0.94$}$ & $61.31\text{\scriptsize$\pm2.29$}$
        & $38.69\text{\scriptsize$\pm1.39$}$ & $38.98\text{\scriptsize$\pm1.53$}$ \\
      & LwF \cite{li2017learning} & 100
        & $49.96\text{\scriptsize$\pm0.68$}$ & $44.22\text{\scriptsize$\pm0.53$}$
        & $40.15\text{\scriptsize$\pm0.43$}$ & $49.61\text{\scriptsize$\pm0.68$}$
        & $30.38\text{\scriptsize$\pm1.41$}$ & $58.15\text{\scriptsize$\pm0.20$}$
        & $45.95\text{\scriptsize$\pm0.72$}$ & $23.21\text{\scriptsize$\pm1.05$}$ \\
      & iCaRL \cite{rebuffi2017icarl} & 100
        & $68.65\text{\scriptsize$\pm0.41$}$ & $31.74\text{\scriptsize$\pm1.89$}$
        & $37.78\text{\scriptsize$\pm2.14$}$ & $55.06\text{\scriptsize$\pm3.50$}$
        & $31.40\text{\scriptsize$\pm1.96$}$ & $59.65\text{\scriptsize$\pm2.40$}$
        & $33.40\text{\scriptsize$\pm0.40$}$ & $53.32\text{\scriptsize$\pm0.90$}$ \\
      \midrule

      \multirow{5}{*}{Prompt-based}
        & L2P \cite{wang2022learning} & 0.2
        & $83.16\text{\scriptsize$\pm0.50$}$ & $10.97\text{\scriptsize$\pm2.02$}$ 
        & $35.26\text{\scriptsize$\pm0.81$}$ & $28.17\text{\scriptsize$\pm0.05$}$
        & $31.65\text{\scriptsize$\pm0.16$}$ & $31.02\text{\scriptsize$\pm1.62$}$
        & $39.39\text{\scriptsize$\pm0.24$}$ & $27.27\text{\scriptsize$\pm1.67$}$ \\
      & Dual-Prompt \cite{wang2022dualprompt} & 0.3
        & $85.47\text{\scriptsize$\pm0.39$}$ & $9.47\text{\scriptsize$\pm1.26$}$
        & $36.59\text{\scriptsize$\pm2.01$}$ & \second{$\mathbf{18.90}\text{\scriptsize$\pm1.63$}$}
        & $33.01\text{\scriptsize$\pm1.65$}$ & $24.61\text{\scriptsize$\pm1.11$}$
        & $39.52\text{\scriptsize$\pm0.33$}$ & $17.13\text{\scriptsize$\pm0.94$}$ \\
      & CODA-Prompt \cite{smith2023coda} & 4.3
        & $83.07\text{\scriptsize$\pm2.56$}$ & $11.04\text{\scriptsize$\pm1.95$}$
        & $38.92\text{\scriptsize$\pm0.27$}$ & $20.55\text{\scriptsize$\pm2.87$}$
        & $31.40\text{\scriptsize$\pm0.36$}$ & $30.54\text{\scriptsize$\pm2.63$}$
        & $42.79\text{\scriptsize$\pm0.51$}$ & $18.63\text{\scriptsize$\pm1.16$}$ \\
      & Sim-CompILer \cite{zhang2024not} & 3.1
        & $84.68\text{\scriptsize$\pm0.58$}$ & $9.98\text{\scriptsize$\pm0.34$}$
        & $42.27\text{\scriptsize$\pm0.63$}$ & $23.89\text{\scriptsize$\pm0.61$}$
        & $33.30\text{\scriptsize$\pm0.10$}$ & $30.31\text{\scriptsize$\pm0.03$}$
        & $41.08\text{\scriptsize$\pm0.27$}$ & $21.20\text{\scriptsize$\pm0.18$}$ \\
      & CompILer \cite{zhang2024not} & 3.1
        & $86.36\text{\scriptsize$\pm0.36$}$ & $7.76\text{\scriptsize$\pm0.68$}$
        & \second{$\mathbf{44.17}\text{\scriptsize$\pm1.41$}$} & $23.28\text{\scriptsize$\pm2.00$}$
        & $34.43\text{\scriptsize$\pm0.07$}$ & $28.69\text{\scriptsize$\pm0.82$}$
        & $42.49\text{\scriptsize$\pm0.35$}$ & $20.36\text{\scriptsize$\pm0.10$}$ \\
      \midrule

      \multirow{4}{*}{Adapter-based}
        & EASE \cite{zhou2024expandable} & 2.2
        & $76.77\text{\scriptsize$\pm0.18$}$ & $9.79\text{\scriptsize$\pm0.18$}$ 
        & $39.58\text{\scriptsize$\pm0.14$}$ & $24.26\text{\scriptsize$\pm0.33$}$ 
        & $34.65\text{\scriptsize$\pm0.22$}$ & \second{$\mathbf{19.60}\text{\scriptsize$\pm0.43$}$}
        & \second{$\mathbf{48.19}\text{\scriptsize$\pm0.15$}$} & $18.79\text{\scriptsize$\pm0.13$}$ \\
      & TUNA \cite{wang2025integrating} & 2.2
        & \second{$\mathbf{89.53}\text{\scriptsize$\pm0.67$}$} & \second{$\mathbf{4.59}\text{\scriptsize$\pm1.22$}$}
        & $43.24\text{\scriptsize$\pm1.38$}$ & $24.09\text{\scriptsize$\pm0.89$}$
        & \best{$\mathbf{40.90}\text{\scriptsize$\pm0.45$}$} & $23.23\text{\scriptsize$\pm0.43$}$
        & $45.08\text{\scriptsize$\pm0.60$}$ & $23.32\text{\scriptsize$\pm0.24$}$ \\
      & CL-LoRA \cite{he2025cl} & 0.5
        & $83.07\text{\scriptsize$\pm0.15$}$ & $6.70\text{\scriptsize$\pm0.51$}$ 
        & $43.00\text{\scriptsize$\pm1.44$}$ & $28.02\text{\scriptsize$\pm1.71$}$ 
        & $37.71\text{\scriptsize$\pm0.80$}$ & $29.25\text{\scriptsize$\pm1.45$}$
        & $36.89\text{\scriptsize$\pm0.09$}$ & \second{$\mathbf{13.58}\text{\scriptsize$\pm1.06$}$} \\
      \cmidrule(lr){2-11}
      & \textbf{LSA-PAST (ours)} & 1.8
        & \best{$\mathbf{91.86}\text{\scriptsize$\pm0.99$}$} & \best{$\mathbf{2.61}\text{\scriptsize$\pm0.32$}$}
        & \best{$\mathbf{44.71}\text{\scriptsize$\pm0.68$}$} & \best{$\mathbf{14.22}\text{\scriptsize$\pm2.18$}$}
        & \second{$\mathbf{39.78}\text{\scriptsize$\pm0.86$}$} & \best{$\mathbf{16.78}\text{\scriptsize$\pm0.57$}$}
        & \best{$\mathbf{48.57}\text{\scriptsize$\pm1.09$}$} & \best{$\mathbf{11.58}\text{\scriptsize$\pm0.46$}$} \\
      \bottomrule
    \end{tabular}%
  }
  \vspace{-1mm}
  \caption{Average accuracy and forgetting on Split-Clothing, Split-UT-Zappos under 5-task and 10-task, and Split-C-GQA. \%Params denotes cumulative additional parameters relative to the ViT-B/16 backbone. \textcolor{bestred}{\textbf{Best}} and \textcolor{secondblue}{\textbf{Second-Best}} results highlighted.}
  \label{tab:avg_acc_ftt}
  \vspace{-4mm}
\end{table*}

\subsection{Optimization Objective}
\noindent \textbf{Training.}
We train with a simple weighted multi-task objective over the three linear classification heads,

\begin{equation}
\mathcal{L} \;=\; \mathcal{L}_{\text{comp}} \;+\; \lambda_{\text{prim}}\big(\mathcal{L}_{\text{state}} + \mathcal{L}_{\text{obj}}\big),
\label{eq:loss_fn}
\end{equation}

\noindent where each loss term $\mathcal{L}_{\text{comp}}$, $\mathcal{L}_{\text{state}}$, and $\mathcal{L}_{\text{obj}}$ is computed as standard cross-entropy for its respective classification head and $\lambda_{\text{prim}}$ is the primitive weight for the state and object losses. Moreover, PAST does not alter the loss function, adding no extra forward-time cost during training.

\noindent \textbf{Inference.}
At task $t$ inference, we discard the linear composition head and classify with cosine similarity to stored prototypes using ensemble over adapter spaces. For each adapter space $a \le t$, the composition branch produces a feature $f_{\text{comp}}^{(a)}(x)$ for an image $x$, which we unit-normalize to obtain $\widehat{z}_{\text{comp}}^{(a)}(x)$. Each composition class $c \in \mathcal{C}^{\le t}$ is first introduced at task $t_c$ and has a native prototype $p_c^{(t_c)}$ in adapter space $t_c$. Across subsequent tasks, we populate a prototype matrix that stores cross-subspace views of each class. When a prototype for class $c$ is available in adapter space $a$, we denote it by $p_c^{(a)}$.  We predict the label for $x$ by ensembling cosine similarities across all adapter spaces up to $t$:

\vspace{-4mm}
\begin{equation}
\widehat{y} \;=\;
\arg\max_{c \in \mathcal{C}^{\le t}}
\sum_{a = 1}^{t}
w_{c,a}\,
\widehat{z}_{\text{comp}}^{(a)}(x)^\top p_c^{(a)},
\end{equation}

\noindent We assign full weight to the native adapter of class $c$ ($w_{c,t_c}=1$) and a smaller constant weight to all other available adapter spaces ($w_{c,a}=0.1$ for $a \neq t_c$ in all experiments). 

\vspace{-3mm}
\section{Experiments and Results}
\label{sec:experiment}
\vspace{-1mm}
\subsection{Experimental Setup}
\noindent \textbf{Datasets.} We evaluate on three composition-IL benchmarks: Split-Clothing \cite{zhang2022learning}, Split-UT-Zappos \cite{yu2014fine}, and our proposed Split-C-GQA dataset derived from C-GQA \cite{naeem2021learning}. Existing work evaluates only on smaller-scale, heavily recurring benchmarks while our Split-C-GQA introduces a larger, less-redundant setting. Split-Clothing has 35 compositions (9 states, 8 objects) across 5 tasks with all primitives recurring, Split-UT-Zappos has 80 compositions (15 states, 12 objects) with 9 states and all objects recurring, Split-C-GQA has 100 compositions (31 states, 31 objects) with 13 states and 18 objects recurring. Detailed statistics are summarized in the supplemental material.

\begin{table}[t!]
  \centering
  \scriptsize 
  \setlength{\tabcolsep}{2pt} 
  \resizebox{1.0\columnwidth}{!}{
    \begin{tabular}{l l cc cc}
      \toprule
      \multirow{2}{*}{Type} & \multirow{2}{*}{Method} 
        & \multicolumn{2}{c}{Split-Clothing} 
        & \multicolumn{2}{c}{Split-C-GQA} \\
      & & State \% & Object \% & State \% & Object \% \\
      \midrule
      \multirow{3}{*}{\rotatebox{90}{Classic}} 
        & EWC \cite{kirkpatrick2017overcoming} 
        & $86.49\text{\scriptsize$\pm0.97$}$ 
        & $52.72\text{\scriptsize$\pm1.30$}$ 
        & $52.91\text{\scriptsize$\pm0.28$}$ 
        & $60.42\text{\scriptsize$\pm1.91$}$ \\
        & LwF \cite{li2017learning} 
        & $87.11\text{\scriptsize$\pm0.66$}$ 
        & $54.57\text{\scriptsize$\pm0.69$}$ 
        & \second{$\mathbf{58.33}\text{\scriptsize$\pm0.42$}$} 
        & $66.32\text{\scriptsize$\pm0.31$}$ \\
        & iCaRL \cite{rebuffi2017icarl} 
        & $91.21\text{\scriptsize$\pm1.05$}$ 
        & $71.70\text{\scriptsize$\pm0.99$}$ 
        & $46.96\text{\scriptsize$\pm0.27$}$ 
        & $55.61\text{\scriptsize$\pm0.45$}$ \\
      \midrule
      \multirow{4}{*}{\rotatebox{90}{Prompt}} 
        & L2P \cite{wang2022learning} 
        & $85.41\text{\scriptsize$\pm0.23$}$ 
        & $95.42\text{\scriptsize$\pm0.42$}$ 
        & $51.65\text{\scriptsize$\pm1.54$}$ 
        & $63.53\text{\scriptsize$\pm1.04$}$ \\
        & Dual-Prompt \cite{wang2022dualprompt} 
        & $86.89\text{\scriptsize$\pm0.37$}$ 
        & $95.87\text{\scriptsize$\pm0.53$}$ 
        & $49.52\text{\scriptsize$\pm0.69$}$ 
        & $64.15\text{\scriptsize$\pm0.37$}$ \\
        & Sim-CompILer \cite{zhang2024not} 
        & $87.14\text{\scriptsize$\pm0.98$}$ 
        & $95.91\text{\scriptsize$\pm0.12$}$ 
        & $51.85\text{\scriptsize$\pm0.45$}$ 
        & $63.89\text{\scriptsize$\pm0.38$}$ \\
        & CompILer \cite{zhang2024not} 
        & $87.76\text{\scriptsize$\pm1.29$}$ 
        & \second{$\mathbf{96.46}\text{\scriptsize$\pm0.08$}$} 
        & $52.70\text{\scriptsize$\pm0.20$}$ 
        & $64.54\text{\scriptsize$\pm0.22$}$ \\
      \midrule
      \multirow{3}{*}{\rotatebox{90}{Adapt}} 
        & EASE \cite{zhou2024expandable} 
        & $77.50\text{\scriptsize$\pm0.16$}$ 
        & $93.73\text{\scriptsize$\pm0.14$}$ 
        & $58.19\text{\scriptsize$\pm0.14$}$ 
        & \second{$\mathbf{69.93}\text{\scriptsize$\pm0.17$}$} \\
        & TUNA \cite{wang2025integrating} 
        & \second{$\mathbf{92.73}\text{\scriptsize$\pm0.93$}$} 
        & $95.84\text{\scriptsize$\pm0.19$}$ 
        & $56.38\text{\scriptsize$\pm0.23$}$ 
        & $65.66\text{\scriptsize$\pm0.09$}$ \\
        & CL-LoRA \cite{he2025cl} 
        & $84.81\text{\scriptsize$\pm0.27$}$ 
        & $96.00\text{\scriptsize$\pm0.11$}$ 
        & $45.26\text{\scriptsize$\pm0.17$}$ 
        & $59.28\text{\scriptsize$\pm0.38$}$ \\
      \midrule
        & \textbf{Ours} 
        & \best{$\mathbf{94.24}\text{\scriptsize$\pm0.21$}$}
        & \best{$\mathbf{96.64}\text{\scriptsize$\pm0.25$}$} 
        & \best{$\mathbf{58.74}\text{\scriptsize$\pm0.28$}$} 
        & \best{$\mathbf{71.01}\text{\scriptsize$\pm0.23$}$} \\
      \bottomrule
    \end{tabular}%
  }
  \caption{State and Object Accuracy (\%) on Split-Clothing and Split-C-GQA (5 tasks). \textcolor{bestred}{\textbf{Best}} and \textcolor{secondblue}{\textbf{Second-Best}} results highlighted.}
  \label{tab:combined_results_small}
  \vspace{-5mm}
\end{table}

\noindent \textbf{Baselines and Metrics.} We opt to compare with Classic CIL \cite{kirkpatrick2017overcoming, li2017learning, rebuffi2017icarl}, prompt-based methods \cite{wang2022learning, wang2022dualprompt, smith2023coda, zhang2024not}, and adapter-based CIL \cite{zhou2024expandable, wang2025integrating, he2025cl}, using LAMDA-PILOT \cite{sun2025pilot, zhou2024class, zhou2024continual} and PyCIL \cite{zhou2023pycil} implementations with tested optimal hyperparameters. iCaRL is the only method which uses exemplars. Results come from \textit{our re-implementations} using the same set of 3 random seeds.

\noindent \textbf{Metrics.} The metrics reported are Average Accuracy (\textbf{Avg. Acc}) over all composition classes seen up to the current task, computed by evaluating the model on every previously seen task after each training step and averaging their accuracies and Forgetting (\textbf{FTT}), which quantifies the degradation on past tasks by comparing each task's current accuracy to its best accuracy observed earlier.

\noindent \textbf{Implementation Details.} We leverage ViT-B/16 \cite{dosovitskiy2020image} with ImageNet-21k pretrain following all prior works \cite{zhou2024expandable,he2025cl,zhang2024not}. The multitask loss is set at $\lambda_{\text{prim}} = 0.1$. We set $(\eta_{\text{both}}, \eta_{\text{one}}) = (0.2, 0.1)$ for Split-Clothing and Split-UT-Zappos, and $(0.15, 0.075)$ for Split-C-GQA. All hyperparameters are selected once using only the first-task split and kept fixed for all subsequent tasks; we do not perform per-task hyperparameter tuning. A detailed complexity analysis, including storage, inference FLOPs, latency, and per-task training time, is provided in the supplemental material.

\subsection{Experimental Results}
\vspace{-1mm}
Table~\ref{tab:avg_acc_ftt} reports Avg Acc and FTT across all benchmarks, including a longer 10-task Split-UT-Zappos setting. LSA-PAST achieves the strongest overall accuracy-forgetting trade-off with low parameter overhead by combining depth-specialized adapters, PGS-anchored prototypes, and AST-based cross-task alignment. On Split-Clothing, LSA-PAST reaches 91.86\% Avg Acc with only 2.61\% FTT, improving over CompILer (86.36\%, 7.76\%) while using fewer trainable parameters. On Split-C-GQA, we also obtain the best Avg Acc and lowest FTT, showing that the method remains effective on a larger benchmark with lower primitive recurrence. On 5-task Split-UT-Zappos, LSA-PAST slightly improves Avg Acc over CompILer (44.71\% vs.\ 44.17\%) but substantially reduces forgetting (14.22\% vs.\ 23.28\%), demonstrating superior stability. Under the longer 10-task schedule, TUNA obtains higher Avg Acc, but LSA-PAST achieves lower FTT (16.78\% vs.\ 23.23\%), demonstrating that our method better preserves prior compositions as task sequences grow. Overall, LSA-PAST delivers the most consistent accuracy-forgetting balance across all settings.

Table~\ref{tab:combined_results_small} reports state and object accuracy on Split-Clothing and Split-C-GQA. LSA-PAST achieves the best primitive-level accuracy across both benchmarks, supporting depth-specialized LoRA for preserving state/object knowledge and reducing state confusion while incrementally learning compositions.

\begin{table}[t!]
  \centering
  \setlength{\tabcolsep}{3.5pt}
  \renewcommand{\arraystretch}{0.8}
  \setlength{\extrarowheight}{1pt}
  \small
  \resizebox{1.0\columnwidth}{!}{%
  \begin{tabular}{cc l|cc|cc}
    \toprule
    \multicolumn{3}{c|}{} &
      \multicolumn{2}{c|}{Split-Clothing} &
      \multicolumn{2}{c}{Split-UT-Zappos} \\
    \cmidrule(lr){4-5}\cmidrule(lr){6-7}

    \textbf{LSA} & \textbf{PGS} & \textbf{Inference}
      & Avg Acc \% ($\uparrow$) & FTT \% ($\downarrow$)
      & Avg Acc \% ($\uparrow$) & FTT \% ($\downarrow$) \\
    \midrule

    \cmark & \cmark & Unaligned Ens.
      & $75.71\text{\scriptsize$\pm0.39$}$
      & $28.32\text{\scriptsize$\pm0.53$}$
      & $29.92\text{\scriptsize$\pm0.37$}$
      & $53.21\text{\scriptsize$\pm1.51$}$ \\

    \cmark & \xmark & Native
      & $90.29\text{\scriptsize$\pm0.65$}$
      & $4.97\text{\scriptsize$\pm0.48$}$
      & $40.85\text{\scriptsize$\pm0.93$}$ & $17.56\text{\scriptsize$\pm1.71$}$ \\

    \cmark & \cmark & Native
      & $90.74\text{\scriptsize$\pm0.65$}$
      & $3.91\text{\scriptsize$\pm0.37$}$
      & $41.67\text{\scriptsize$\pm0.57$}$
      & $14.81\text{\scriptsize$\pm1.85$}$ \\

    \cmark & \xmark & \textbf{AST Ens. (ours)}
      & $90.81\text{\scriptsize$\pm0.27$}$
      & $4.89\text{\scriptsize$\pm0.87$}$
      & $44.58\text{\scriptsize$\pm1.56$}$
      & $18.63\text{\scriptsize$\pm3.17$}$ \\

    \midrule

    \cmark & \cmark & LDC Ens. \cite{gomez2024exemplar}
      & $91.41\text{\scriptsize$\pm0.40$}$
      & $3.49\text{\scriptsize$\pm0.73$}$
      & $43.73\text{\scriptsize$\pm0.77$}$
      & $15.09\text{\scriptsize$\pm1.74$}$ \\

    \cmark & \cmark & DPCR Ens. \cite{he2025semantic}
      & $91.25\text{\scriptsize$\pm0.84$}$
      & $3.72\text{\scriptsize$\pm0.58$}$
      & $44.07\text{\scriptsize$\pm0.74$}$
      & $15.26\text{\scriptsize$\pm1.98$}$ \\

    \cmark & \cmark & \textbf{AST Ens. (ours)}
      & $\mathbf{91.86}\text{\scriptsize$\pm0.99$}$
      & $\mathbf{2.61}\text{\scriptsize$\pm0.32$}$
      & $\mathbf{44.71}\text{\scriptsize$\pm0.68$}$
      & $\mathbf{14.22}\text{\scriptsize$\pm2.18$}$ \\

    \bottomrule
  \end{tabular}%
  }
  \caption{Component and inference ablation. Native scores each class only in its originating adapter space, while ensemble (Ens.) variants additionally aggregate available cross-space evidence.}
  \label{tab:pgs_ablate}
  \vspace{-5mm}
\end{table}

\begin{figure*}[t!]
    \centering
    \includegraphics[width=0.75\textwidth]{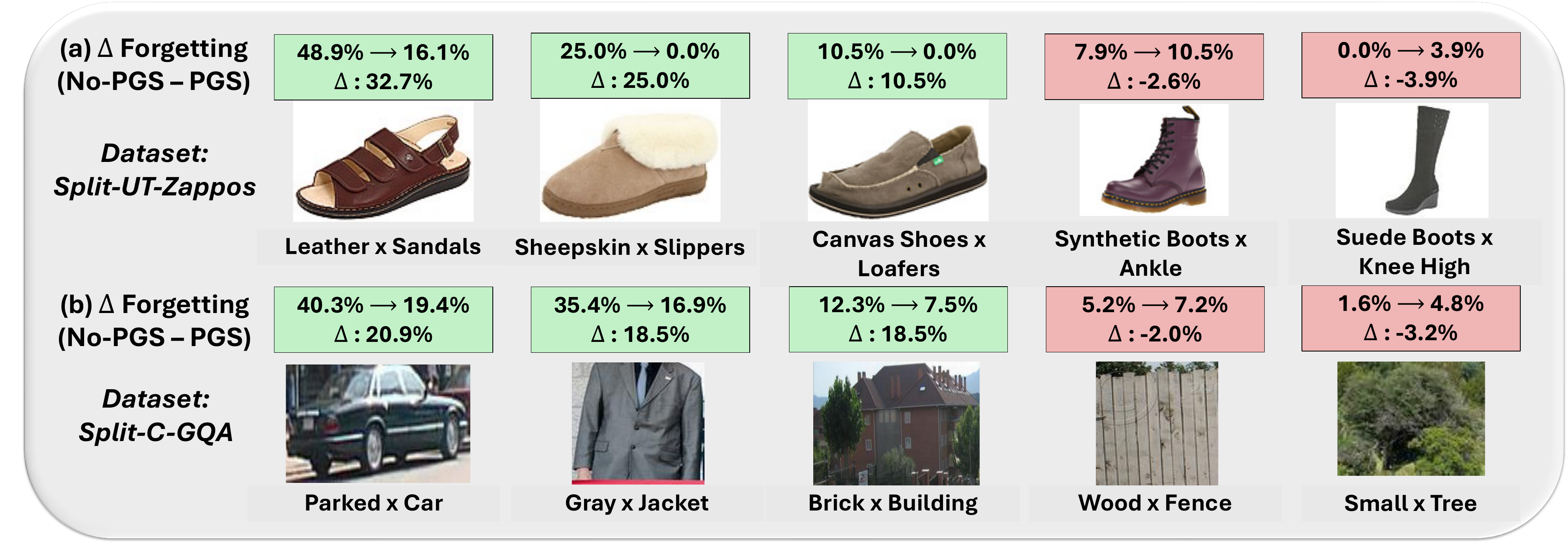}
    \caption{PGS success and failure cases on Split-UT-Zappos and Split-C-GQA. Boxes show $\Delta$ forgetting, with green indicating reduced forgetting and red indicating no improvement.}
    \label{fig:Qualitative}
    \vspace{-3mm}
\end{figure*}

\vspace{-1mm}
\subsection{Ablation Study}
\vspace{-1mm}
\noindent\textbf{Component and Transport Ablation.}
We ablate the components of LSA-PAST in Table \ref{tab:pgs_ablate}. The unaligned ensemble performs poorly, confirming that scores cannot be reliably aggregated across task-specific adapter spaces without alignment. Native-space inference provides a strong transport-free baseline by evaluating each class only in its originating space. Adding PGS consistently reduces forgetting in both Native and AST inference, supporting its role in stabilizing composition prototypes through recurring primitive knowledge. With PGS fixed, AST improves over Native by retaining reliable native evidence while incorporating lower-weight aligned evidence from non-native spaces, increasing Avg.\ Acc.\ from $90.74\%$ to $91.86\%$ on Split-Clothing and from $41.67\%$ to $44.71\%$ on Split-UT-Zappos. Finally, under the same LSA+PGS backbone and ensemble protocol, AST outperforms previous drift compensation methods LDC \cite{gomez2024exemplar} and DPCR \cite{he2025semantic}, yielding the strongest overall accuracy-forgetting trade-off.


\noindent\textbf{PGS Qualitative Analysis.}
Fig.~\ref{fig:Qualitative} shows representative PGS success and failure cases on Split-UT-Zappos and Split-C-GQA. PGS substantially reduces forgetting for visually distinctive states such as ``Leather'' and ``Parked,'' where primitive information provides a reliable direction for prototype stabilization. In contrast, ambiguous states such as ``Small'' or ``Synthetic'' can produce slight degradation, suggesting that primitive anchoring is less reliable when state evidence is weak or visually subtle.

\noindent\textbf{Oracle Prototype Alignment.}
For each past class, we construct an \textit{oracle prototype} from held-out test images passed through the current adapter and compare it with the Uncorrected prototype from task $t'$, LDC-corrected~\cite{gomez2024exemplar}, DPCR-corrected \cite{he2025semantic}, and AST-transported prototypes using cosine similarity (Table \ref{tab:oracle}). Results are grouped by task gap $t-t'$, while Overall averages all past-to-current task-pair comparisons. LDC and DPCR are slightly stronger at gap 1 on Split-Clothing and Split-C-GQA, respectively, but AST performs best at gaps 2-4 and achieves the highest overall alignment on both datasets. This demonstrates that AST more closely approximates the current-space location of old-class prototypes as task separation increases.

\vspace{-1mm}
\begin{table}[t!]
\centering
\setlength{\tabcolsep}{2.9pt}
\renewcommand{\arraystretch}{0.9}
\large
\resizebox{1.0\columnwidth}{!}{%
\begin{tabular}{lcccc|cccc}
\toprule
& \multicolumn{4}{c|}{Split-Clothing}
& \multicolumn{4}{c}{Split-C-GQA} \\
\cmidrule(lr){2-5}\cmidrule(lr){6-9}

\textbf{Task Gap}
& Uncorrected & LDC \cite{gomez2024exemplar} & DPCR \cite{he2025semantic} & AST
& Uncorrected & LDC \cite{gomez2024exemplar} & DPCR \cite{he2025semantic} & AST \\
\midrule

1
& $0.740$ & $\mathbf{0.898}$ & $0.863$ & $0.880$
& $0.784$ & $0.942$ & $\mathbf{0.945}$ & $0.941$ \\

2
& $0.750$ & $0.873$ & $0.830$ & $\mathbf{0.897}$
& $0.786$ & $0.928$ & $0.934$ & $\mathbf{0.945}$ \\

3
& $0.704$ & $0.844$ & $0.812$ & $\mathbf{0.886}$
& $0.798$ & $0.910$ & $0.920$ & $\mathbf{0.948}$ \\

4
& $0.662$ & $0.819$ & $0.799$ & $\mathbf{0.905}$
& $0.774$ & $0.881$ & $0.894$ & $\mathbf{0.952}$ \\

\midrule
\textbf{Overall}
& $0.728$ & $0.872$ & $0.836$ & $\mathbf{0.889}$
& $0.786$ & $0.925$ & $0.932$ & $\mathbf{0.944}$ \\

\bottomrule
\end{tabular}%
}
\caption{Cosine similarity to oracle current-space prototypes.}
\label{tab:oracle}
\vspace{-5mm}
\end{table}

\begin{figure}[t!]
\vspace*{-5mm}
\centering
\includegraphics[width=1.0\columnwidth]{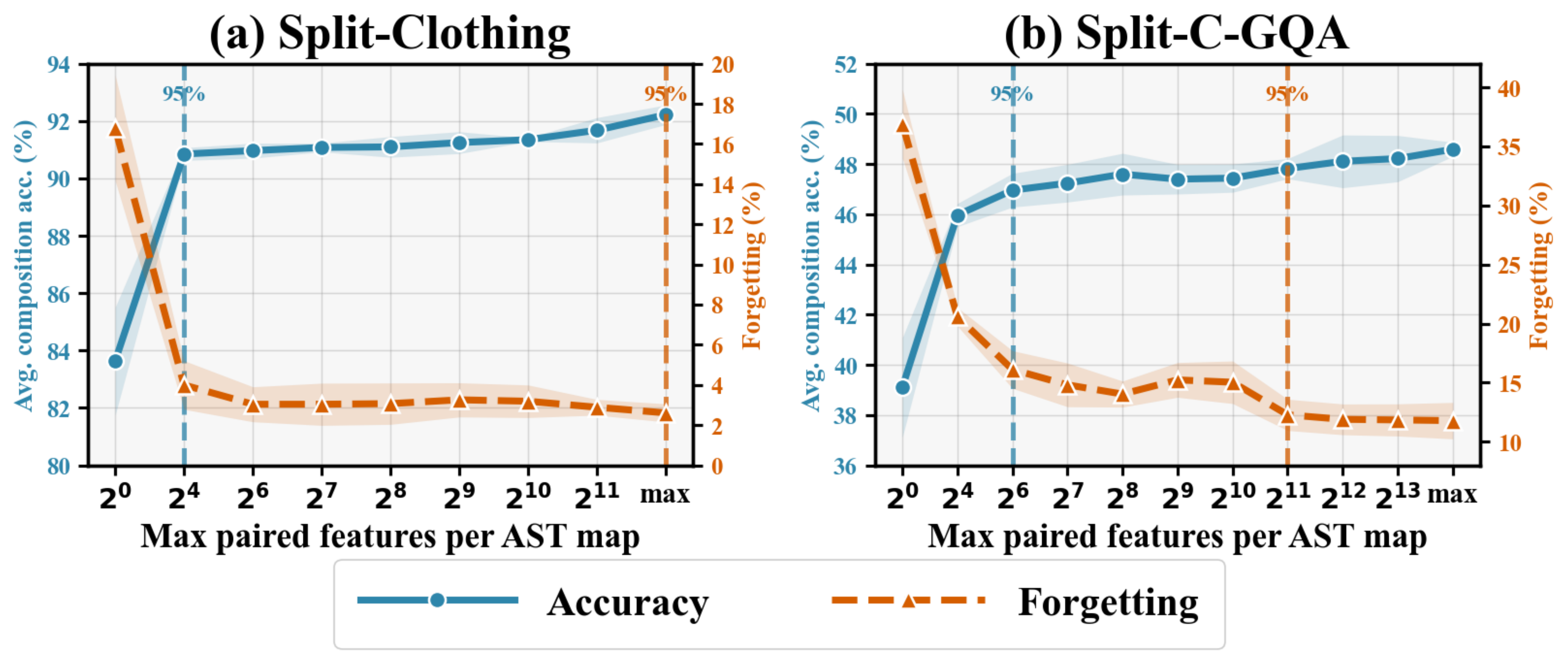}

\vspace{-2mm}

\includegraphics[width=1.0\columnwidth]{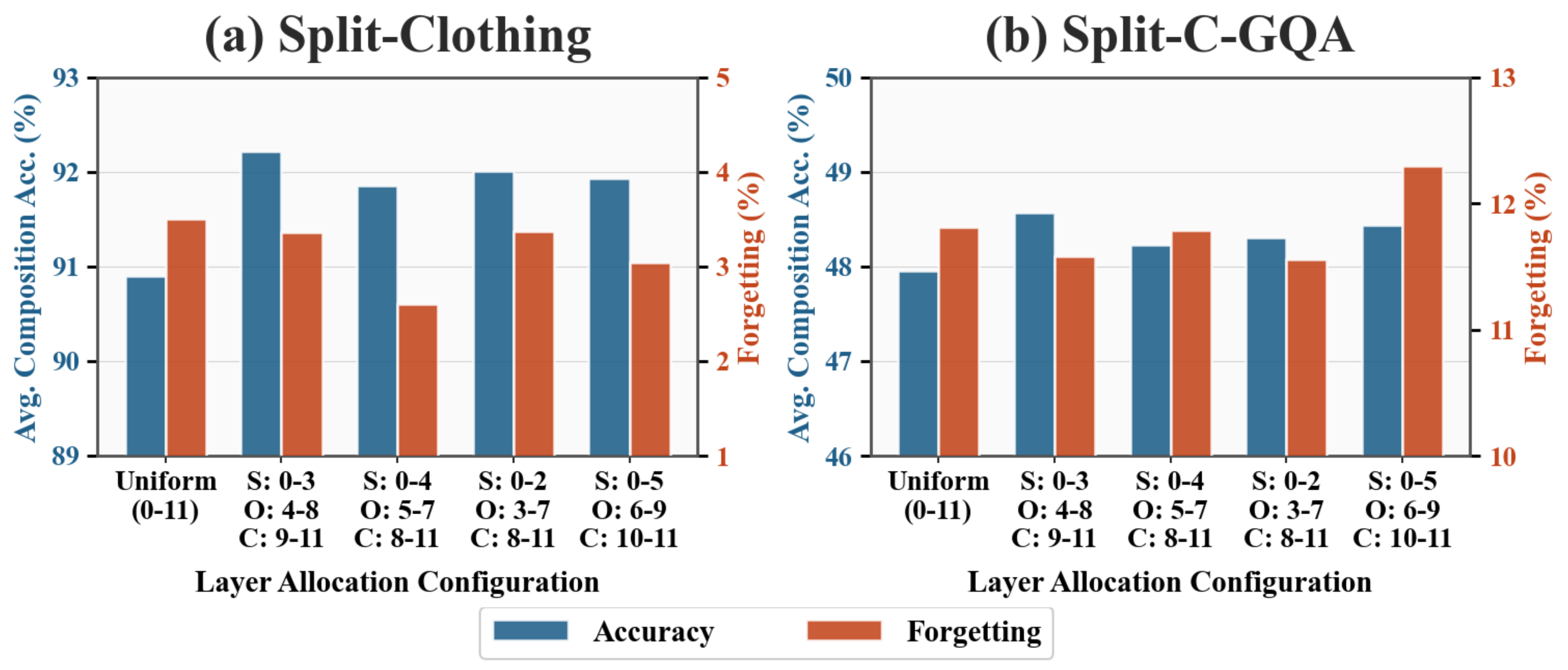}

\vspace{-2mm}
\caption{Top: effect of AST transport-pair budget on accuracy and forgetting. Bottom: LSA ablation with explicit state (S), object (O), and composition (C) layer ranges.}
\label{fig:ast_lsa_ablation}
\vspace{-4mm}
\end{figure}

\vspace{1mm}
\noindent\textbf{Effect of Number of Transported Pairs.}
The top panel of Fig.~\ref{fig:ast_lsa_ablation} varies the number of current-task feature correspondences used to fit each AST map. Accuracy improves rapidly with a small transport budget, while additional correspondences further reduce forgetting by refining cross-space alignment. Increasing the budget affects only post-task AST fitting and does not increase inference-time cost. The 95\% markers show that near-maximum accuracy is reached with fewer correspondences than near-optimal forgetting, suggesting that coarse alignment is sufficient for accuracy gains, whereas stronger retention benefits from finer transport estimates.

\noindent\textbf{Effect of Layer Assignment.} The bottom panel of Fig.~\ref{fig:ast_lsa_ablation} compares depth-specialized LSA with alternative layer assignments, including uniform LoRA under a matched parameter budget, where identical adapters are attached to all layers for S/O/C. Our LSA assignment follows the \textit{approximate} hierarchy observed in prior work \cite{dorszewski2025colors, NEURIPS2024_6216515a, lepori2024beyond}: state cues are assigned to early layers, object structure to middle layers, and global operation to the final layers. While the uniform design is competitive, LSA consistently improves Avg Acc and further reduces FTT using the same number of trainable parameters. Depth specialization aligns adaptation with this hierarchy, avoiding uniform updates that can \textit{entangle} unrelated features. This design also enables depthwise control (e.g., different LoRA ranks per depth), offering a more flexible alternative to uniform LoRA.

\vspace{-2mm}
\section{Conclusion}
\label{sec:conclusion}
\vspace{-1mm}


In this paper, we propose LSA-PAST, a framework for composition-IL that exploits the hierarchical processing structure of vision transformers and a novel, rehearsal-free cross-subspace alignment method for aligning old prototypes with current task subspaces. Through extensive experiments across multiple benchmark datasets, we demonstrate that our method achieves SOA performance in accuracy-forgetting trade off. Comprehensive studies validate the importance of both depth-specialized adapter placement and AST to overall performance. Finally, we provide contributions to an emerging field of incremental learning, introducing a framework that lays foundational groundwork for future research in compositional continual models.

{
    \small
    \bibliographystyle{ieeenat_fullname}
    \bibliography{main}
}

\end{document}